\documentclass{article}

 \usepackage[preprint]{neurips_2026}

\usepackage[utf8]{inputenc} 
\usepackage[T1]{fontenc}    
\usepackage{hyperref}       
\usepackage{url}            
\usepackage{booktabs}       
\usepackage{amsfonts}       
\usepackage{nicefrac}       
\usepackage{microtype}      
\usepackage{xcolor}         

\usepackage{graphicx}
\usepackage{amsmath}
\usepackage{subcaption}
\usepackage{enumitem}
\usepackage{multirow}

\title{Detect in Any Scene: An Agentic Framework for Object Detection with Experience-Aware Reasoning}

\author{%
  Wenlun Zhang \\
  Keio University\\
  \texttt{wenlun\_zhang@keio.jp} \\
  \And
  Jun Yin \\
  Tsinghua University\\
  \texttt{yinj24@mails.tsinghua.edu.cn} \\
  \And
  Kentaro Yoshioka \\
  Keio University\\
  \texttt{kyoshioka47@keio.jp} \\
}

\begin{document}

\maketitle

\begin{abstract}
  Object detection in real-world scenarios remains challenging due to diverse image degradations and heterogeneous object distributions, which significantly hinder the generalization of existing detectors. Conventional approaches, including scene-specific representation learning and end-to-end pipeline design, are inherently limited by their reliance on predefined conditions and lack adaptability to dynamic environments. In this paper, we propose DetAS, an agentic detection framework that formulates object detection as a dynamic decision process. Instead of relying on static pipelines, DetAS leverages a Multimodal Large Language Model (MLLM) as a central agent to adaptively compose detection workflows by selecting from a toolbox of restoration modules and specialized detectors. Specifically, DetAS consists of two key components: Self-Adaptive Image Restoration, which dynamically determines whether and how to enhance images for downstream detection, and Multi-Expertise Detection, which integrates multiple domain-specialized detectors and resolves their predictions through instance-level reasoning. To further improve decision quality under fine-grained conditions, we introduce Self-Evolving Experience Harvesting and extend the framework to DetAS-X, which accumulates node-level decision experience from a small set of annotated data and enables experience-aware reasoning during inference. This mechanism allows the system to progressively refine its decision policy and adapt to diverse real-world scenarios. Extensive experiments on six challenging benchmarks demonstrate that DetAS-X significantly outperforms existing MLLM-based detectors, achieving an average improvement of 28.36\% in F1 score, with up to 37.01\% gain on DarkFace. These results demonstrate the promise of agentic detection and establish a solid foundation for its application in complex and dynamic environments.
  \textbf{Code will be open-sourced upon acceptance.}
\end{abstract}

\section{Introduction}

Object detection has achieved remarkable progress~\cite{SSD,DETR,DINO}, yet existing models often struggle to generalize to real-world scenarios. In practice, images are frequently affected by diverse degradations, such as haze, rain, low-light conditions, and underwater distortion, which significantly alter visual appearance and degrade detection performance. At the same time, real-world applications involve highly diverse object categories and scene distributions, further increasing task complexity. Consequently, detectors trained under standard settings struggle to transfer across different degradation types and application scenarios~\cite{SHIFT}.

\begin{figure}[t!]
    \centering
    \begin{subfigure}{0.32\linewidth}
        \centering
        \includegraphics[width=\linewidth]{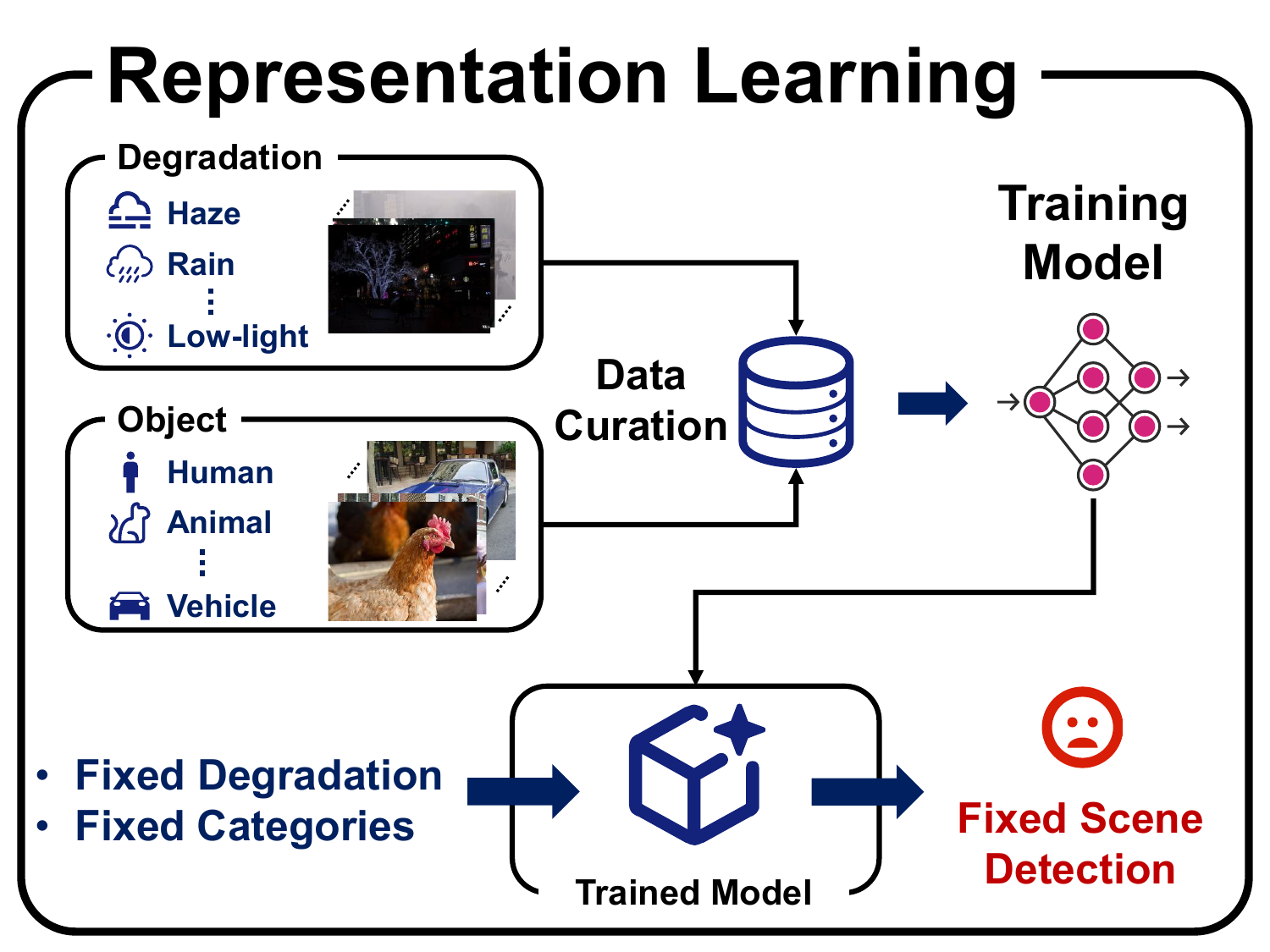}
        \caption{}
    \end{subfigure}
    \hfill
    \begin{subfigure}{0.32\linewidth}
        \centering
        \includegraphics[width=\linewidth]{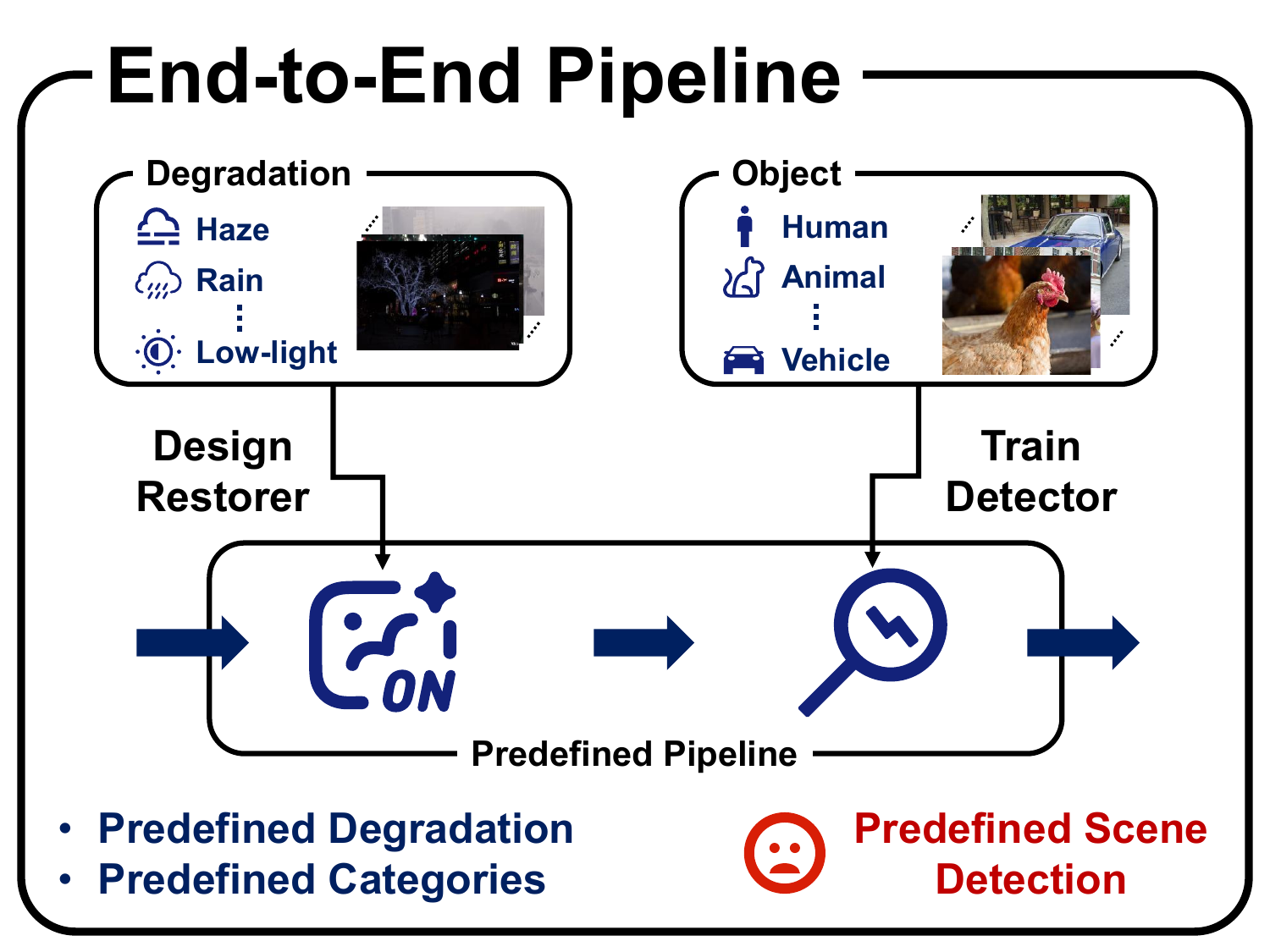}
        \caption{}
    \end{subfigure}
    \hfill
    \begin{subfigure}{0.32\linewidth}
        \centering
        \includegraphics[width=\linewidth]{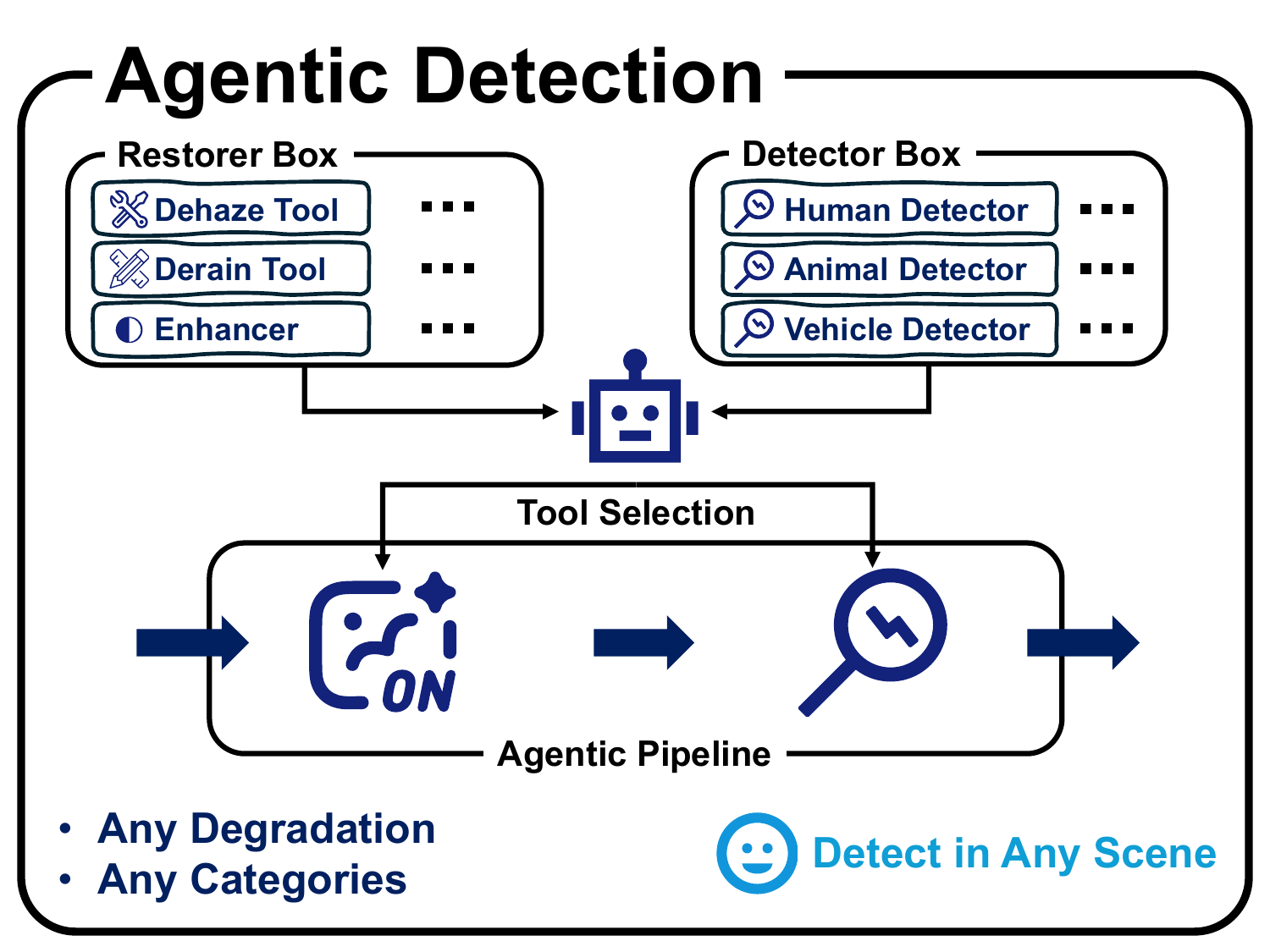}
        \caption{}
    \end{subfigure}
    \caption{Comparison of object detection paradigms: (a) representation learning with curated datasets, (b) end-to-end pipeline design, and (c) our proposed agentic detection framework that dynamically selects tools for adaptive detection across diverse scenarios.}
    \label{Fig_Overview}
\end{figure}

Existing approaches to robust object detection under degraded conditions can be broadly categorized into two paradigms: representation learning and end-to-end pipeline. \textbf{Representation Learning} methods rely on curated datasets with fixed degradation types and object categories to learn specialized representations~\cite{ACDC,FoggyCityscapes}, as shown in Fig.~\ref{Fig_Overview}(a). While effective within the training domain, these methods often fail to generalize under even slight shifts in degradation or scene distribution, leading to significant performance degradation. Alternatively, \textbf{End-to-End Pipeline} adopts a modular approach~\cite{IA-YOLO,DID-MDN}, where a detector trained under standard conditions is combined with a dedicated image restoration module for specific scenarios, as shown in Fig.~\ref{Fig_Overview}(b). Although this design avoids large-scale data collection and retraining, it typically relies on predefined restoration strategies and manually designed pipelines. As a result, suboptimal restoration may degrade detection performance, and the entire pipeline often requires redesign when adapting to new scenarios. Therefore, both paradigms are inherently limited to fixed degradation conditions and detection targets, lacking the flexibility and transferability required for diverse and dynamic environments. In particular, traditional closed-vocabulary detectors struggle to generalize in dynamically changing real-world scenarios, where a single detector or fixed pipeline cannot simultaneously adapt to varying environmental conditions and object distributions.

The limitations of existing paradigms suggest that robustness under diverse degradations requires adaptive, input-aware detection rather than improved representations or fixed pipelines. Recent advances in Multimodal Large Language Models (MLLMs)~\cite{MLLM_Survey}, particularly in grounding and visual reasoning~\cite{MiMo-VL,Seed1.5-VL}, along with the emergence of agentic systems~\cite{Toolformer,4KAgent}, provide a compelling foundation for this paradigm shift. Large Language Model (LLM) agents demonstrate strong capabilities in understanding complex visual scenes, reasoning over multiple candidates, and making context-aware decisions. Such properties naturally align with the requirements of real-world object detection, where both degradation conditions and target objects vary significantly across samples.

Building on this perspective, we propose an \textbf{Agentic Detection} framework that formulates object detection as a dynamic decision-making process, as illustrated in Fig.~\ref{Fig_Overview}(c). Instead of treating detection as a static pipeline, our approach automatically composes a workflow by selecting from a diverse toolbox of image restoration modules and specialized detectors. Given an input image, the system first performs scene perception and then adaptively selects appropriate restoration strategies and detection modules. Based on this idea, we develop \underline{Det}ect in \underline{A}ny \underline{S}cene (DetAS), which consists of Self-Adaptive Image Restoration (SAIR) and Multi-Expertise Detection (MED). As shown in Fig.~\ref{Fig_Framework}, SAIR dynamically selects restoration strategies through iterative LLM perception, while MED integrates multiple specialized detectors to collaboratively handle diverse detection scenarios. To further improve reasoning under fine-grained conditions, we introduce Self-Evolving Experience Harvesting (SEEH), resulting in an upgraded framework, DetAS-X. DetAS-X harvests node-level decision experience from a small set of annotated data and leverages this experience to enable experience-aware reasoning during inference, leading to improved detection accuracy and a self-evolving detection system that adapts to diverse real-world scenarios. Our contributions are summarized as follows:

\begin{itemize}[leftmargin=*]
    \item We propose an agentic detection framework, termed DetAS, which consists of SAIR and MED for adaptive selection of restoration tools and detectors.
    \item We further introduce SEEH and extend our framework to DetAS-X, which leverages experience from annotated data to enable experience-aware reasoning and improve detection performance.
    \item Extensive experiments demonstrate that DetAS-X consistently outperforms existing MLLM baselines and generalizes well across diverse degradations and scenarios. In particular, DetAS-X achieves an average improvement on F1 score of 28.36\% across six datasets, with up to 37.01\% improvement on DarkFace under low-light face detection.
\end{itemize}

\section{Related Works}

\textbf{Robust Object Detection under Degradation:} Object detection under degraded visual conditions has progressed along two main directions. The first constructs degradation-specific datasets and learns specialized representations. Benchmarks such as DarkFace~\cite{DarkFace} highlight the performance gap of standard detectors under low-light conditions, while datasets like BDD100K~\cite{BDD100K} and HazyDet~\cite{HazyDet} expand coverage to diverse weather conditions and viewpoints. Methods built on these datasets learn degradation-aware representations or adaptation strategies, such as HLA-Face~\cite{HLA-Face}, AERIS~\cite{AERIS}, and DAI-Net~\cite{DAI-Net}. However, these approaches are typically tied to specific degradation types or domains and generalize poorly across scenarios. The second direction develops restoration-assisted detection pipelines. Early work such as AOD-Net~\cite{AOD-Net} couples dehazing with detection, and recent methods (e.g., VRD-IR~\cite{VRD-IR}, SR4IR~\cite{SR4IR}, GEFU~\cite{GEFU}, UniRestore~\cite{UniRestore}) explicitly optimize restoration for downstream recognition. Despite their effectiveness, these approaches typically rely on fixed restoration modules and detectors, and seldom consider whether restoration is necessary or how to select the optimal restoration–detector combination for a given input. Motivated by this limitation, we move beyond static pipelines and propose a more adaptive detection framework.

\textbf{Open-Vocabulary Visual Grounding and Detection:} Visual grounding and open-vocabulary detection underpin language-conditioned object localization, typically achieved through vision–language alignment. Methods such as GLIP~\cite{GLIP} and Grounding DINO~\cite{Grounding_DINO} integrate large-scale language supervision into detection frameworks, while MLLMs including Qwen3-VL~\cite{Qwen3-VL}, GLM-4.1V~\cite{GLM-4.1V}, and InternVL3.5~\cite{InternVL3.5} demonstrate strong grounding capabilities for open-world understanding. Rex-Omni~\cite{Rex-Omni} further improves detection, particularly for small and dense objects. More recent work extends grounding from final prediction to intermediate reasoning. For example, Rex-Thinker~\cite{Rex-Thinker} introduces Chain-of-Thought (CoT) reasoning, P2G~\cite{P2G} dynamically invokes external modules, GroundingAgent~\cite{GroundingAgent} employs agentic reasoning, and VGR~\cite{VGR} enhances reasoning via region selection and reuse. Despite these advances, existing methods are primarily evaluated on clean and static scenarios, leaving robustness under severe degradation, occlusion, and dynamic conditions insufficiently explored.

\section{Methodology}

\begin{figure*}[t!]
    \centering
    \includegraphics[width=\textwidth]{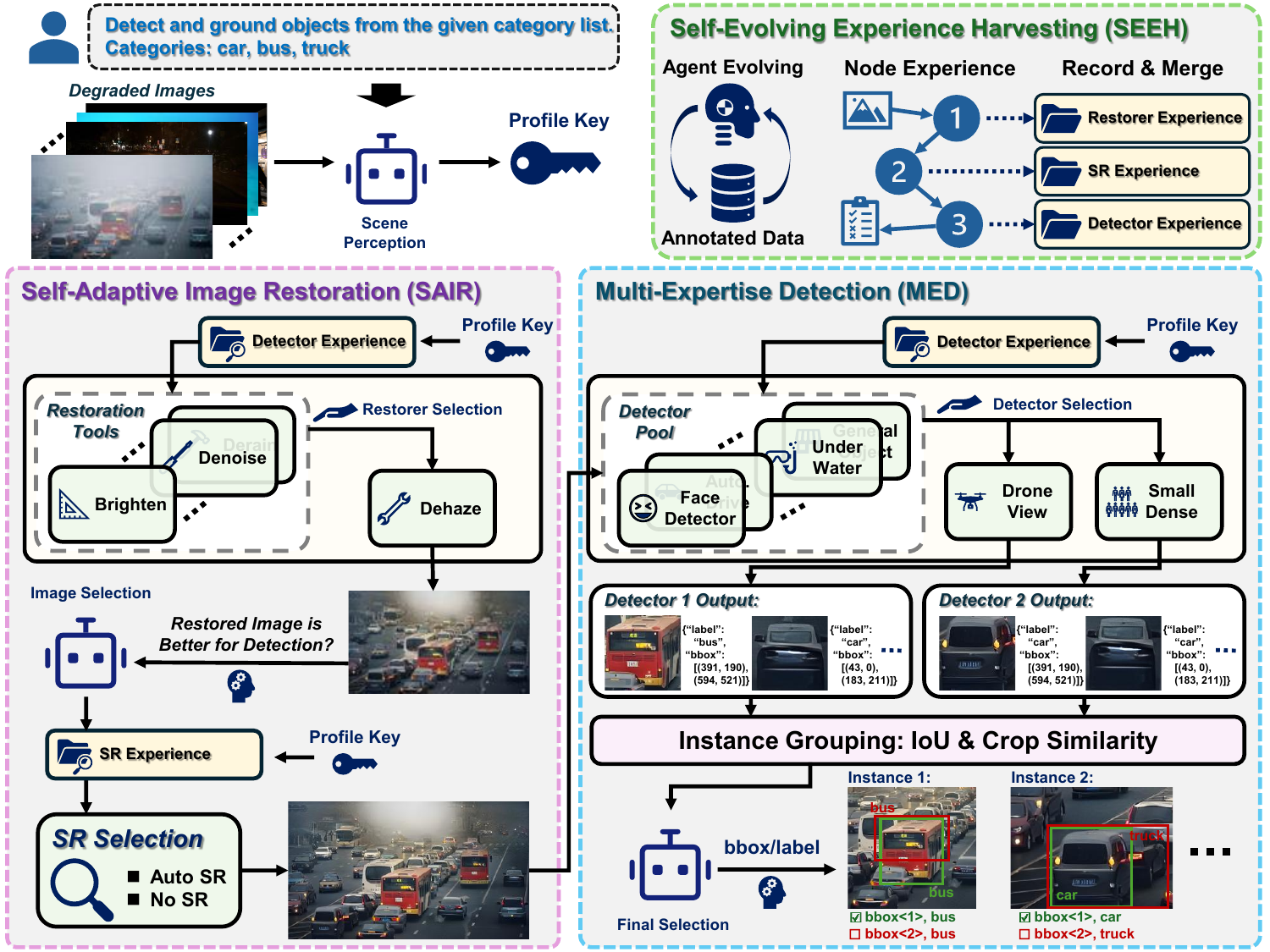}
    \caption{Overview of the DetAS-X framework. SAIR adaptively selects restoration strategies through LLM perception, while MED integrates multiple specialized detectors to handle diverse detection scenarios. SEEH harvests node-level decision experience from annotated data to enable experience-aware reasoning and improve detection performance.}
    \label{Fig_Framework}
\end{figure*}

To build an agentic detection system for diverse degradations and scenarios, we propose DetAS, a framework that automatically selects restoration modules and specialized detectors to construct effective pipelines. As shown in Fig.~\ref{Fig_Framework}, DetAS includes two key components: SAIR, which produces detection-friendly images, and MED, which leverages complementary detectors for diverse scenarios. To further improve decision quality, we introduce SEEH and extend the framework to DetAS-X, which accumulates experience from annotated data to progressively enhance detection performance.

\subsection{Self-Adaptive Image Restoration}

Images in real-world environments are often affected by diverse degradations, which can significantly impair object detection performance. To address this issue, as shown in Fig.~\ref{Fig_Framework}, DetAS introduces a SAIR pipeline that dynamically selects appropriate restoration strategies based on the input image condition. Specifically, an LLM perception module first analyzes the image and predicts a degradation-aware scene profile from a predefined set, including \texttt{normal}, \texttt{fog}, \texttt{rain}, \texttt{underwater}, \texttt{low-light}, and \texttt{noise}. Based on the predicted scene label, SAIR automatically activates the corresponding restoration tool from a restorer pool that includes dehazing, deraining, denoising, brightness enhancement, and other restoration tools.

However, restoration does not always improve downstream detection performance. In some cases, restoration may alter background colors or distort object boundaries, which can negatively affect detection accuracy. To address this issue, SAIR introduces an image selection mechanism that compares the original image with the restored version and selects the one that is more suitable for detection. If restoration degrades the image quality for detection, the original image is used instead. Importantly, the selection decision prioritizes detection-oriented cues such as object visibility, boundary clarity, and structural integrity rather than visual aesthetics, ensuring that the chosen image is more suitable for downstream object detection.

In the final restoration stage, SAIR applies Super-Resolution (SR) to increase spatial resolution and improve the visibility of target objects. This step is particularly useful when small or distant objects are present in the scene. Given a predefined target resolution, the SR module automatically selects an appropriate upscaling factor to enhance image clarity.

Overall, through this adaptive restoration pipeline, SAIR produces a detection-friendly image representation that enhances object visibility and provides improved input representation for downstream detection modules.

\subsection{Multi-Expertise Detection}

Instead of relying on a single detector, DetAS adopts a MED pipeline that integrates multiple specialized detectors to handle diverse object distributions across a wide range of scenarios. During scene perception, the LLM also receives a target category list. By jointly considering the image content and target categories, the LLM-based agent selects the most suitable top-$K$ detectors from a detector pool consisting of MLLM detectors with diverse expertise, including \texttt{general-purpose detector}, \texttt{dense\&small-object detector}, \texttt{autonomous-driving detector}, \texttt{drone-view detector}, \texttt{underwater detector}, and \texttt{face detector}. Each detector specializes in particular visual patterns or object distributions, enabling the system to handle a wide range of detection scenarios and generate comprehensive initial grounding proposals.

After detector selection, the chosen detectors are applied to the preprocessed image to generate candidate object proposals, as shown in Fig.~\ref{Fig_Framework}. Since multiple detectors may produce duplicate predictions for the same object instance, DetAS performs an instance grouping step to consolidate detection results. Candidate bounding boxes are grouped based on spatial overlap and visual similarity between grounded regions, forming instance groups that likely correspond to the same object. Formally, let $\mathcal{B}=\{b_n\}_{n=1}^{N}$ denote the set of candidate bounding boxes predicted by multiple detectors. We first compute the Intersection-over-Union (IoU) between two candidate boxes $b_i$ and $b_j$. If the IoU exceeds a predefined threshold, the two proposals are considered spatially consistent. However, spatial overlap alone is insufficient when detectors produce slightly shifted predictions, especially when objects appear in dense scenes. To improve grouping robustness, we additionally compute a visual similarity score between their cropped image regions. For each candidate box $b_i$, the bounding box is expanded with ratio $\alpha$ to obtain an enlarged crop region, which is subsequently resized to $32\times32$ patch and flattened into a vector representation. The visual similarity between two cropped regions is defined as:
\begin{equation}
S_{\text{vis}}(i,j) = \frac{\mathbf{v}_i^\top \mathbf{v}_j}{\|\mathbf{v}_i\|_2 \|\mathbf{v}_j\|_2},
\label{Eq_Visual_Similarity}
\end{equation}
where $\mathbf{v}_i$ denotes the vectorized representation of the cropped region. This lightweight similarity computation avoids the overhead of using a vision encoder to extract latent embeddings, thereby maintaining high efficiency. Finally, a candidate proposal $b_i$ is assigned to an existing group $\mathcal{G}=\{g_k\}_{k=1}^{K}$ if both spatial and visual consistency are satisfied:
\begin{equation}
\text{IoU}(b_i,b_j) > \theta_{\text{IoU}} \quad \land \quad S_{\text{vis}}(i,j) > \phi_{\text{vis}},
\label{Eq_Instance_Grouping}
\end{equation}
where $\theta_{\text{IoU}}$ and $\phi_{\text{vis}}$ denote the IoU and visual similarity thresholds, respectively.

In the final stage of MED, the system selects the best bounding box–label pair for each instance group. For each group, we extract the region defined by the grouped bounding boxes and provide it, together with the corresponding detection proposals, to the LLM for final decision making. Compared with prompting on the entire image, instance-level crops are more token-efficient and encourage the LLM to focus on the target object. This design is particularly beneficial for small objects, where the LLM may otherwise fail to attend to relevant regions and produce suboptimal decisions. For illusory false prediction groups, where detectors fail to identify valid objects or repeatedly generate background proposals, the system allows rejecting all candidates within the group. Overall, this MED strategy enables DetAS to leverage complementary detector strengths while relying on LLM reasoning to resolve ambiguous predictions.

\subsection{Self-Evolving Experience Harvesting}

To this end, DetAS establishes a rule-based agentic detection flow that performs effectively across diverse scenarios. However, decisions based solely on predefined rules lack the flexibility to capture fine-grained characteristics of individual images. In practice, images with similar degradation types may still require different restoration or detection strategies. For example, applying dehazing to lightly foggy images may blur object features, while SR may amplify restoration artifacts and confuse downstream detectors. In such cases, bypassing the corresponding restoration or SR step and directly performing detection on the original image can yield better results. Moreover, the optimal combination of detectors often varies across scenarios, highlighting the need for more fine-grained detector selection. To address these challenges, we introduce SEEH, which enables the system to automatically collect, store, and summarize node-level decision experience from a small set of annotated data. The accumulated experience is then leveraged to guide the agent’s decisions, allowing the detection system to progressively evolve and improve performance in real-world applications.

As shown in Fig.~\ref{Fig_Framework}, we introduce an experience retrieval mechanism for three critical decision nodes in DetAS: restorer selection, SR selection, and detector selection, resulting in an experience-aware framework, DetAS-X. Before deployment, we collect a small set of annotated images and perform exploratory runs to obtain decision knowledge via empirical evaluation. For each image, the LLM-based agent first infers a fine-grained scene profile during perception, including the scene label as well as attributes such as \texttt{object scale}, \texttt{object density}, \texttt{visibility}, and \texttt{illumination}. This profile serves as a key for organizing and retrieving experience. During experience harvesting, the system executes the DetAS pipeline on annotated images under different decision configurations (right-top panel in Fig.~\ref{Fig_Framework}), exploring restoration strategies, SR options, and detector selections. Detection performance is evaluated using ground truth, and for each decision node, the experience manager records metrics such as F1 scores. For images sharing the same profile, records are aggregated based on performance statistics to identify the most effective tools, forming a compact experience memory for guiding future decisions. Unlike many memory-based methods that summarize experience as trajectories or textual descriptions, SEEH directly records detection results at each decision node. This design avoids long contextual prompts that may distract the LLM, providing a concise and effective reference for reliable decision-making.

During inference, DetAS-X incorporates experience-aware reasoning into the inference process. For each input image, the system first performs scene perception to obtain a fine-grained profile key describing its visual characteristics. The experience memory is queried to retrieve the top-$K$ most similar profiles together with their associated recommendations at each decision node. These retrieved experiences are provided to the LLM-based agent as reference information, allowing the system to jointly consider current visual evidence and accumulated historical experience when selecting restoration strategies, SR configurations, and detectors. As more annotated data are processed, the experience memory is continuously expanded and refined to better reflect the deployment environment, enabling DetAS-X to progressively improve its decision policy. Through this self-evolving mechanism, SEEH transforms the detection framework from a rule-based pipeline into an experience-guided adaptive system that continuously learns from past observations and improves performance across diverse real-world scenarios.

\section{Experiments}

\subsection{Experimental Setup}

\textbf{Datasets and Baselines:} We evaluate our methods on multiple benchmarks covering challenging real-world image degradations. Specifically, we include \textbf{HazyDet}~\cite{HazyDet}, a foggy-scene detection dataset captured from drone viewpoints; \textbf{MARIS}~\cite{MARIS} for underwater object detection; \textbf{DarkFace}~\cite{DarkFace} for low-light face detection; and the Night (\textbf{B-Night}) and Rainy (\textbf{B-Rainy}) subsets of BDD100K~\cite{BDD100K} for nighttime and rainy driving environments. We additionally report results on \textbf{COCO}~\cite{COCO} as a reference for standard object detection. We compare our methods with several recent MLLMs for grounding-based object detection. For open-source models under 10B parameters, we include \textbf{Ovis2.5-9B}~\cite{Ovis2.5}, \textbf{Qwen3.5-9B}~\cite{Qwen3.5}, \textbf{MiniCPM-V-4.5}~\cite{MiniCPM-V4.5}, \textbf{GLM-4.1V-9B}~\cite{GLM-4.1V}, \textbf{InternVL3.5-8B}~\cite{InternVL3.5}, and \textbf{Qwen3-VL-8B}~\cite{Qwen3}. We also evaluate two closed-source frontier models, \textbf{GPT-4.1}~\cite{Gpt-4} and \textbf{Gemini-2.5-Flash}~\cite{Gemini}, to provide a comprehensive performance reference. For open-source MLLMs, we perform grounding supervised fine-tuning (SFT) on the training split of each dataset using LoRA with rank 8 for 1 epoch. For closed-source models, we specify the output format and prompt the models directly via their APIs.

\textbf{Implementation Details:} For restoration, we adopt several fast and effective models, including RIDCP~\cite{RIDCP} for dehazing, MPRNet~\cite{MPRNet} for deraining, SwinIR~\cite{SwinIR} for denoising, LLFlow~\cite{LLFlow} for brightness enhancement, and ESRGAN~\cite{ESRGAN} for SR. To construct the detector pool with diverse expertise, we perform SFT of Qwen3-VL-8B on different datasets. Specifically, the \texttt{general-purpose detector} is finetuned on COCO, the \texttt{autonomous-driving detector} on BDD100K, the \texttt{drone-view detector} on clean images before fog synthesis in HazyDet, the \texttt{face detector} on DarkFace, and the \texttt{underwater detector} on MARIS. For \texttt{small\&dense-object detector}, we directly use Rex-Omni~\cite{Rex-Omni} due to its strong performance on dense object scenarios. For perception and reasoning, we employ the original Qwen3-VL-8B instruct model as the LLM-based agent. During instance grouping, the bounding-box expansion ratio $\alpha$ is set to 0.25, while the spatial and visual thresholds ${\theta}_{\text{IoU}}$ and ${\phi}_{\text{vis}}$ in Eq.~\ref{Eq_Instance_Grouping} are set to 0.5. The top-$K$ value for detector selection is set to 2, and the target SR resolution is 2K. For DetAS-X, we first harvest experience on 50 samples randomly drawn from the training split with ground-truth annotations for each dataset. During inference, the top-$K$ value for experience retrieval is set to 3. 

\subsection{Experimental Results}

\textbf{Benchmark Results:} We evaluate detection performance using the F1 score (\%) at $\text{IoU}=0.5$, and report the results in Table~\ref{Table_Experimental_Results}. For CoT-based MLLMs, we disable the thinking mode, as we observe that CoT reasoning often collapses in object detection tasks. Despite being fine-tuned on their respective datasets, baseline MLLMs still exhibit limited detection performance. As shown in Table~\ref{Table_Experimental_Results}, most models achieve low F1 scores on challenging datasets, especially under foggy, low-light, and rainy conditions. For instance, Qwen3.5-9B achieves only 3.25\% on HazyDet, 0.29\% on DarkFace, and 2.00\% on B-Rainy. Even the strongest grounding model, Qwen3-VL-8B, achieves only 31.34\% on HazyDet, 7.97\% on DarkFace, and 3.96\% on B-Rainy. These results indicate that SFT alone is insufficient to bridge the domain gap caused by severe degradations, highlighting the need for adaptive restoration and expertise-aware detection strategies.

In contrast, DetAS significantly improves performance across all datasets through the cooperation of SAIR and MED. For example, on HazyDet, DarkFace, and B-Rainy, DetAS improves the F1 score from 31.34\% to 44.78\%, from 7.97\% to 43.92\%, and from 11.29\% to 40.08\%, respectively, demonstrating the effectiveness of dynamically selecting restoration strategies. On COCO, where input images typically do not require restoration, adaptive selection of specialized detectors still yields substantial improvement, increasing the F1 score from 33.25\% to 59.64\%. Furthermore, with experience-aware reasoning by SEEH, DetAS-X consistently improves upon DetAS, achieving additional gains such as 7.57\% on HazyDet and 6.13\% on MARIS. These results indicate that leveraging historical experience enables more fine-grained tool selection beyond rule-based decisions. Overall, the superior and consistent performance of DetAS-X across all datasets highlights its strong generalization capability, validating the importance of adaptive restoration, multi-expertise collaboration, and experience-aware reasoning for robust object detection in dynamic scenarios.

\begin{table}[t!]
\small
\centering
\caption{Evaluation results on diverse datasets (F1 score (\%) at $\text{IoU}=0.5$). The best results are shown in bold and the second-best are underlined. DetAS-X achieves the best detection performance.}
\label{Table_Experimental_Results}
\begin{tabular}{l|cccccc|c}
\toprule
\textbf{Method} & \textbf{COCO} & \textbf{HazyDet} & \textbf{MARIS} & \textbf{DarkFace} & \textbf{B-Night} & \textbf{B-Rainy} & \textbf{Avg.} \\
\midrule
\textbf{GPT-4.1} & 9.05 & 2.52 & 30.72 & 0.00 & 0.70 & 2.27 & 7.54 \\
\textbf{Gemini-2.5-Flash} & 13.03 & 13.59 & 29.54 & 0.50 & 17.47 & 10.96 & 14.18 \\
\midrule
\textbf{Ovis2.5-9B} & 18.45 & 17.20 & 15.87 & 5.60 & 7.40 & 11.29 & 12.64 \\
\textbf{Qwen3.5-9B} & 23.85 & 3.25 & 48.14 & 0.29 & 0.09 & 2.00 & 12.94 \\
\textbf{MiniCPM-V-4.5} & 17.78 & 12.00 & 36.66 & 0.94 & 5.66 & 5.18 & 13.04 \\
\textbf{GLM-4.1V-9B} & 30.11 & 14.35 & 38.41 & 0.00 & 5.48 & 2.83 & 15.20 \\
\textbf{InternVL3.5-8B} & 33.25 & 19.53 & 24.30 & 6.28 & 7.45 & 7.44 & 16.38 \\
\textbf{Qwen3-VL-8B} & 14.33 & 31.34 & \underline{51.56} & 7.97 & 8.60 & 3.96 & 19.63 \\
\midrule
\textbf{DetAS} & \underline{59.64} & \underline{44.78} & 46.16 & \underline{43.92} & \underline{36.81} & \underline{39.69} & \underline{45.17} \\
\textbf{DetAS-X} & \textbf{59.90} & \textbf{52.35} & \textbf{52.29} & \textbf{44.98} & \textbf{38.35} & \textbf{40.08} & \textbf{47.99} \\
\bottomrule
\end{tabular}
\end{table}

\begin{figure*}[htbp]
    \centering
    \includegraphics[width=\textwidth]{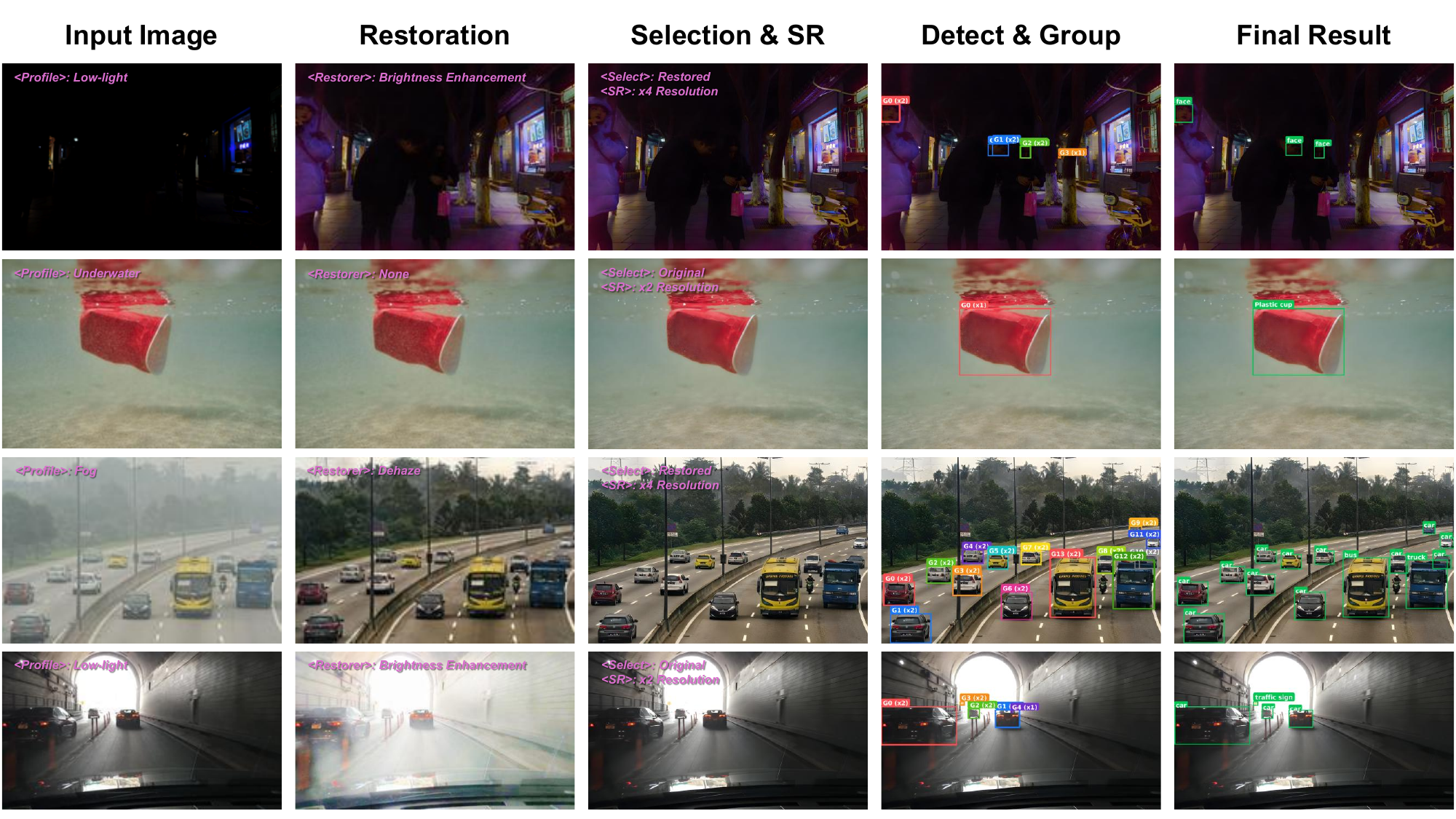}
    \caption{Qualitative visualization of the DetAS-X pipeline under diverse scenarios.}
    \label{Fig_Qualitative_Visualization}
\end{figure*}

\textbf{Qualitative Analysis:} We provide qualitative visualizations of the DetAS-X pipeline to illustrate how the system improves detection under different scenarios (Fig.~\ref{Fig_Qualitative_Visualization}). Each row shows one example and each column corresponds to a stage of the pipeline. In the first example, the scene is identified as low-light, and brightness enhancement followed by SR significantly improves the visibility of the face region, enabling correct detection through the collaboration of multiple detectors and LLM reasoning. For the second example, the system determines that restoration is unnecessary and directly applies SR, after which the underwater-specialized detector produces the final detection result. In the third example, the image is recognized as a foggy scene; dehazing and SR substantially improve vehicle visibility, allowing the LLM to select tighter bounding boxes from multiple detector proposals. For the last example, although restoration is initially recommended, the restored image degrades detection quality, and the LLM therefore selects the original image for SR, which leads to correct detection results. Overall, these examples demonstrate that DetAS-X can adaptively generate detection-friendly images across diverse scenarios and effectively combine multi-detector proposals with LLM reasoning to produce accurate final detections.

\subsection{Ablation Studies}

\begin{figure}[htbp]
    \centering
    \begin{subfigure}{0.49\linewidth}
        \centering
        \includegraphics[width=\linewidth]{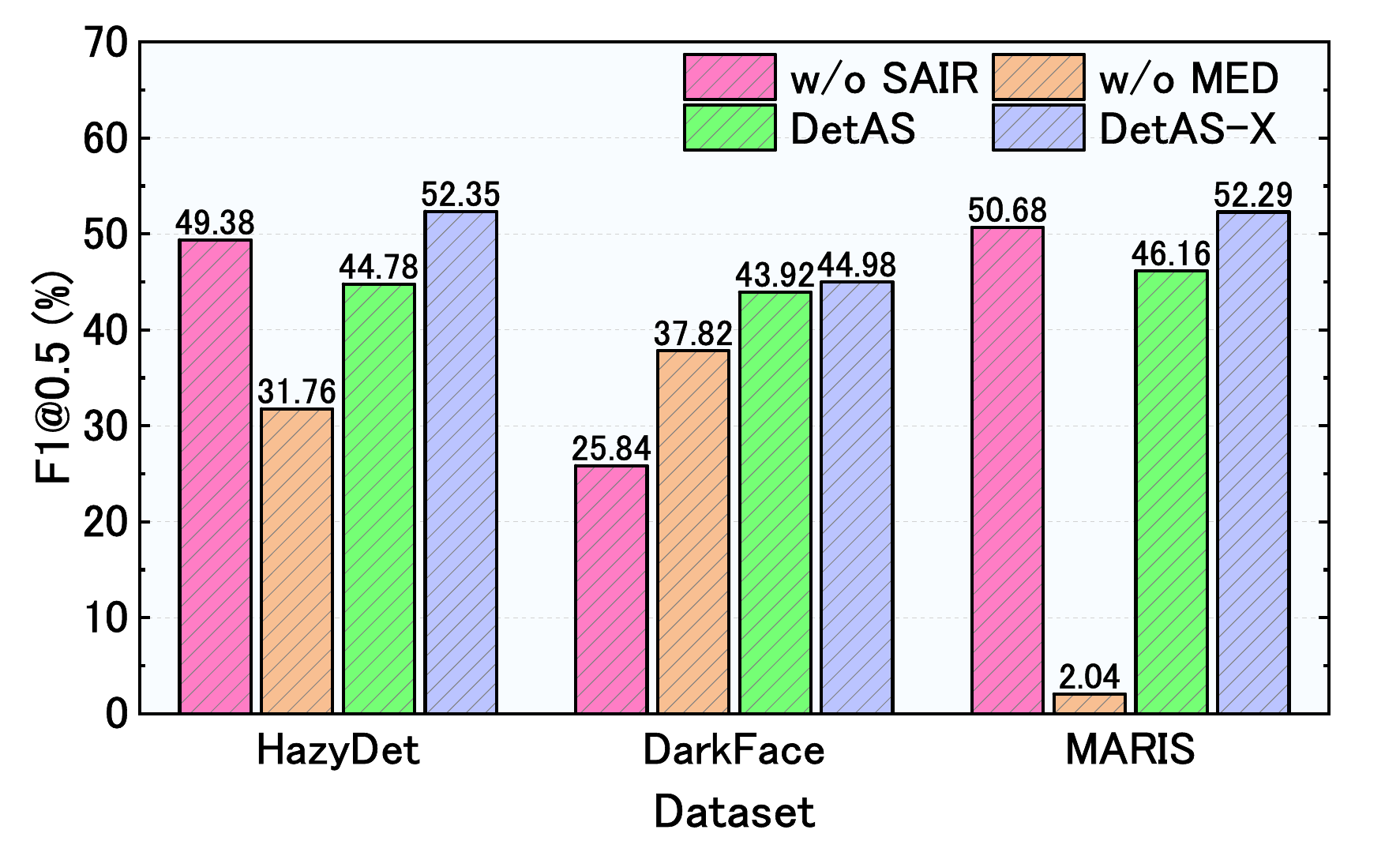}
        \caption{}
    \end{subfigure}
    \hfill
    \begin{subfigure}{0.49\linewidth}
        \centering
        \includegraphics[width=\linewidth]{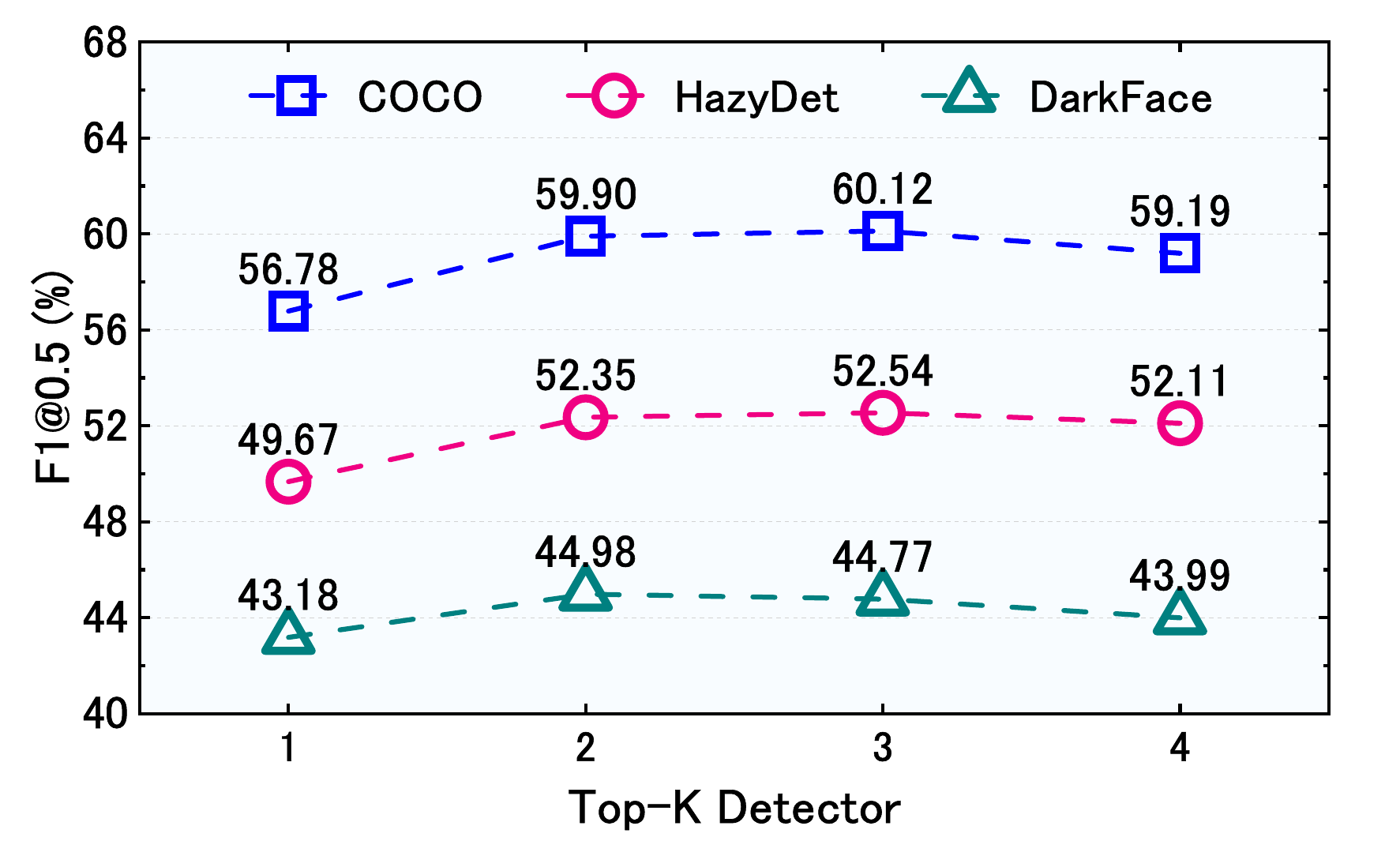}
        \caption{}
    \end{subfigure}
    \caption{Ablation studies: (a) Component contributions. (b) Effect of detector count.}
    \label{Fig_Abaltion}
\end{figure}

We conduct ablation studies on multiple datasets to evaluate the contribution of each module in DetAS-X and the impact of the number of detectors in MED. Fig.~\ref{Fig_Abaltion}(a) presents the component analysis, where we compare DetAS-X with DetAS and two additional ablated variants. To evaluate the effectiveness of SAIR, we remove the SAIR module and directly perform detection on the original input image (denoted as \textbf{w/o SAIR}). To assess the role of MED, we replace it with a \texttt{general-purpose detector} (denoted as \textbf{w/o MED}). The results show that each module consistently contributes to performance across scenarios. In particular, on DarkFace and MARIS, the F1 score drops from 44.98\% to 25.84\% without SAIR and from 52.29\% to 2.04\% without MED, respectively, demonstrating the importance of both modules. Furthermore, incorporating SEEH refines the original rule-based pipeline with accumulated experience, leading to consistent performance improvements. 

We further analyze the impact of the number of selected detectors (Top-$K$) in MED on detection performance. We evaluate configurations with $K=1$ to $4$ across three datasets. Results show that even with a single detector, the LLM-based agent can select a suitable model and achieve reasonable performance. Increasing the number of complementary detectors generally improves performance, but the gain saturates at $K=2$, while using $K=4$ leads to slight degradation. This is likely because out-of-domain detectors tend to introduce false predictions, which negatively affect performance. Therefore, to balance performance and computational cost, we set $K=2$ in the DetAS-X framework. 

\subsection{Visualization of Detection Performance}

\begin{figure*}[htbp]
    \centering
    \includegraphics[width=\textwidth]{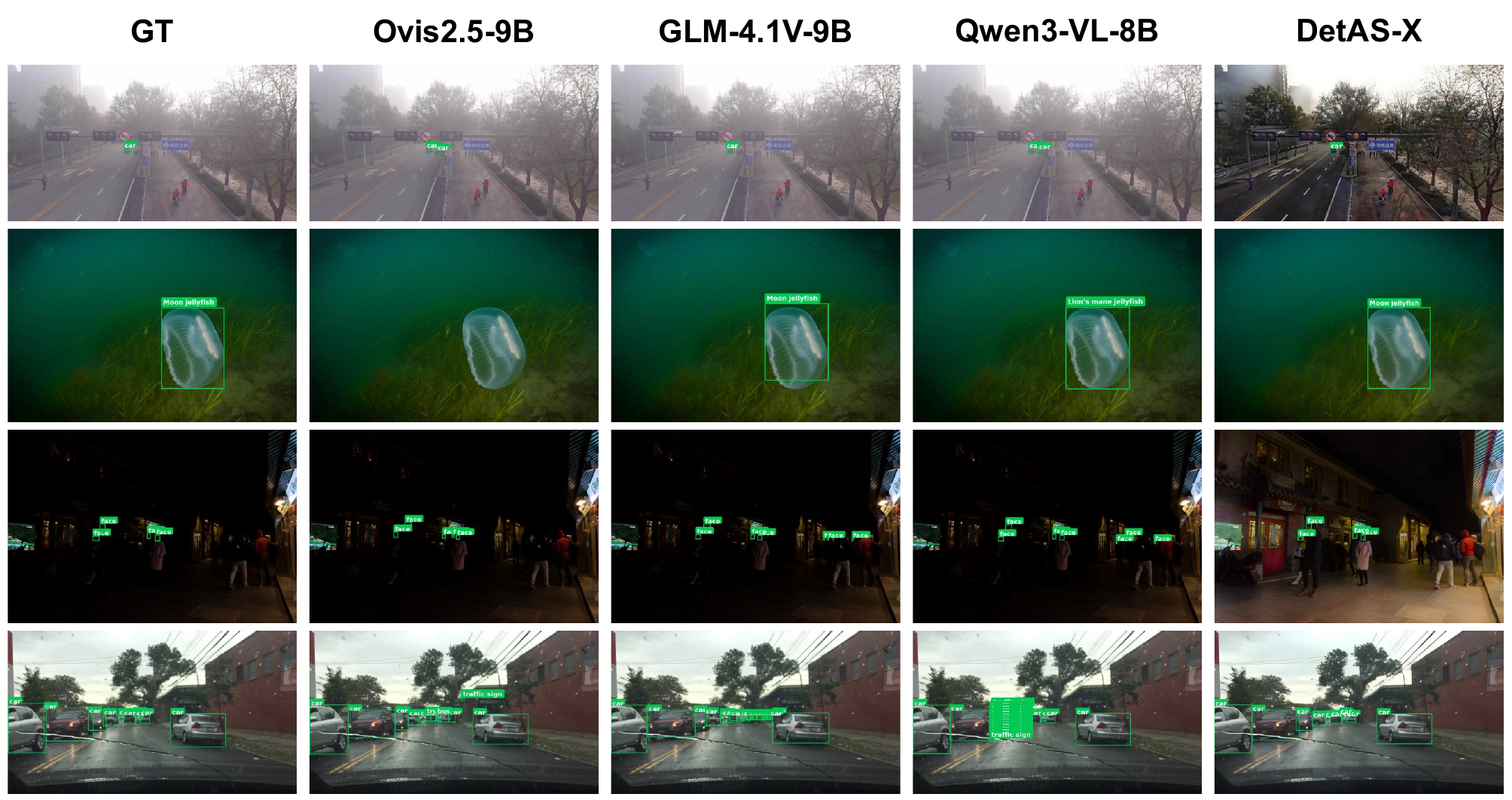}
    \caption{Qualitative comparison of detection results between DetAS-X and baseline MLLMs.}
    \label{Fig_Detection_Visualization}
\end{figure*}

We provide qualitative comparisons of detection results across different baselines in Fig.~\ref{Fig_Detection_Visualization}. Compared with baseline models such as Ovis2.5-9B, GLM-4.1V-9B, and Qwen3-VL-8B, DetAS-X produces more accurate and complete detections under diverse degradation conditions. Baseline methods often suffer from missed detections, imprecise localization, incorrect labels, and excessive false positives in challenging scenarios such as fog, underwater, and low-light environments. In contrast, DetAS-X consistently identifies target objects with tighter bounding boxes and more accurate predictions. These qualitative results further validate the robustness and generalization of DetAS-X in complex real-world settings. 

\section{Limitations}

We discuss several limitations of DetAS-X to highlight future directions. First, the framework is constrained by the capability of MLLM-based detectors, which generally underperform conventional detectors in fixed-scene settings. However, MLLM-based detectors offer open-vocabulary detection capabilities and strong transferability under domain shifts, which have attracted increasing attention from both industry and academia. As MLLM grounding capabilities continue to improve, this limitation is expected to diminish. Second, DetAS-X is constructed using five restoration tools and six domain-specific detectors covering common open-environment scenarios. More specialized domains, such as infrared, X-ray, and medical imaging, are not explored in this work. Nevertheless, the framework is modular and scalable, and can be extended by incorporating additional tools and detectors to support broader applications.

\section{Conclusion}

We propose DetAS, an agentic framework that formulates object detection as a dynamic decision process, enabling adaptive composition of restoration and detection pipelines for diverse real-world scenarios. By integrating SAIR and MED, DetAS overcomes the limitations of static pipelines through dynamic restoration selection and complementary detector expertise. We further introduce SEEH and extend the framework to DetAS-X, which enables experience-aware reasoning by accumulating node-level decision knowledge from annotated data, allowing progressive refinement under fine-grained scene variations. Extensive experiments show that DetAS-X consistently outperforms existing MLLM-based detectors, achieving substantial F1 improvements and strong generalization across diverse degradations. Overall, this work highlights agentic detection as a promising paradigm for robust visual understanding and scalable real-world deployment.



\bibliographystyle{plainnat}
\bibliography{reference}


\end{document}